# Data formats for phonological corpora

Laurent Romary, INRIA & HUB-IDSL

Andreas Witt, Institut für Deutsche Sprache

## Representing annotated spoken corpora

The annotation of linguistic resources has long-standing traditions (see Cole et al., 2010). The other chapters of this book make clear that the production of annotated resources is a laborious, time-consuming, andexpensive task. In theory, we want to make these resources available in such a way that they can be re-used by as many scholars as possible (see Ide&Romary, 2002). However, a largevariety of annotation formatshave been developed in the previous decades, each one created for a specific research task. Consequently,the resulting resources are frequently only usable by members of the individual research projects.

The goal of the present chapter is to explore the possibility of providing the research and industrial communities that commonly use spoken corpora with a set of well-documented standardised formats that allow a high re-use rate of annotated spoken resources and, as a consequence, better interoperability across tools used to produce or exploit such resources. We hope to identify standards thatcoverall possible aspects of the management workflow of spoken data, from the actual representation of raw recordings and transcriptions to high-level content-related information at a semantic or pragmatic level. Most of the challenges here are similar to those for textual resources, except for, on the one hand, the grounding relation that spoken data has to illocutionary circumstances (time, place, speakers and addressees), and, on the other hand, the specific annotation layers that correspond to speech related information (e.g. prosody), comprising multimodal aspects such as gestures.

We should also not forget, as is well illustrated in this book, the importance of legacy practices in the spoken corpora community, most of them resulting from the existence of specific tools at various representation layers, ranging from basic transcription tools (Transcriber, PRAAT) to generic score-based annotation environments (TASX, Elan, CLAN/CHAT (CHILDES), EMU). By definition, these various tools do not have the same maintenance rate and capacity and it is therefore essential to think about standardised formats as offering the possibility to be embedded with existing practices. This implies that we have two basic scenarios in mind:

- We want to be able to project existing data into a range of standardised representations that bear as little specificity to the original format as possible but as much faithfulness as necessary;
- We want standardised formats to havethe capacity to be used for the development of new technical platforms, thus allowing the integration of new requirements and new features.

These two general requirements both imply standards that can incorporate features and data we have not yet envisioned. To do this, the standards should provide specification or customisation mechanisms that do not hinder their abilityto improve interoperability.

That said, it is clear that such a thorough set of standardscannot be fully describedina single book chapter. Moreover, we acknowledge that there is still some work to be done before we will have a convincing portfolio of standards that cancover all aspects of annotated spoken corpora. For these reasons, we are adopting an intentionally selective (and hence subjective) strategy, with the goal of laying out a foundation that can serve as a basis to complete the standardisation picture step by step.

After a brief introduction[1] to existing standardisation activities for language resources in general, we will provide some basic concepts related to the representation of annotated linguistic content. We will present in detail some of the proposals that may be used for the transcription and annotation of spoken data, along with the possibility of defining precise semantics for the corresponding representations.

[1] For a precise presentation of background activities which lead to the current standardization picture, see (Ide and Romary 2007)

## Standards and standardisation processes

It has become common tospeak of two kinds of standards:*de facto* standards, whicharisethrough the practices of active communities and are adopted over the years, and*de jure* standards, which are created “from scratch” and promulgated by official standardisation bodies. Such a dichotomy is misleading, since the actual development of standards is usually accomplished by cooperation from both of these sides. Indeed, we suggest that standardisation isa process with three essential components:

- *Consensus building* within a technical community, including the involvement of reliable experts and the consideration of existing practices and developments;
- The *wide availability* of the standard so that any potential user may determine how much he or she is complying to it;
- A *maintenance process*, through which existing defects or necessary improvements maybe implemented in further revisionsof the standard, while taking care of backward compatibility issues.

These processes are the basis for most standardisation bodies, including official national andinternational organisations such as ISO or IETF, or consortium based bodies such as the W3C, OASIS or the TEI. Many standard proposals that do not arise from these processes (usually those initiated within dedicated research and development projects) have failed or suffered due to the lack of communitysupport that could provide for dissemination and maintenance of the standards.

For language resources, we can identify three main organisations thatplay the most important role in standards:

- The World Wide Web Consortium (W3C)provides horizontal standards (called recommendations) for the management of Internet-based communication, and in particular XML technologies[2], which are widely used for representing all sorts of semi-structured information. The W3C

[2] In the remaining text of this paper, we assume that the reader has some basic understanding of XML technologies, and in particular have no difficulty in reading through the XML samples we introduce. See also (Bray et alii 1998)

also carries out language-oriented activities regarding internationalisation, in particular;

- The International Organisation for Standardisation (ISO), a confederation of national standardisation bodies that covers nearly all areasof industrial activities. Beyond generic IT-relevant projects carried out in ISO-IEC JTC1 (from character encoding with ISO 10646-Unicode to document representation with SGML), technical committee 37 (TC 37) of ISO provides guidance for linguistic content management. In particular, sub-committee 2 (SC 2) of TC 37 is in charge of language codes, SC 3 of computer based terminologies and SC 4 of language resources;
- The Text Encoding Initiative (TEI), a consortium that has taken up the responsibility of offering the digital humanities community at large with a wide range of XML-based representations covering most of the possible useful genres from prose text to dictionaries.

## Which standards for linguistic annotation of spoken corpora?

To understand standards for language resources, it is important to understand the various activities that the standardisation organisations mentioned above are pursuing. In the following paragraphs, we suggest a possible overall strategy to achieve the best standard-based approach to the management of linguistic data and justify the biased approach taken up in the rest of the paper.

### Various user scenarios - various standards

It is important to consider how standardisation relates to possible organisation levels of spoken corpora. In general, these organisation levels include:

- The first important level of representation in phonological corpora is the transcription, where the source signal, in the form of an audio or video file, as well as any additionalinformation provided by specific sensors (e.g. articulatory) is segmented and classified as a set of symbolic codes. Such codes may be phonetic or orthographic ones, but may also correspond to any kind of features or patterns that are deemed useful for further analysis of the primary source. Transcription is understood as a process which theoretically should be independent of further annotation steps;

- Anchored to the transcription layers (also referred to as tiers), but also to other prior annotations, a given annotation layer is identified as providing a certain type of interpretation of the primary source, whether this is linguistic (e.g. the identification of syntactic constructs) or of any other possible kind (e.g. identification of pathological features in the speaker's voice). As we shall see, the specification of an annotation level relies on the provision of its internal logic (meta-model) and the corresponding elementary descriptors (data categories);
- Finally, an important aspect of corpus annotation relies on the proper management of the combination of annotation layers (also called *tiers* in phonological corpora), as well as the corpus of primary sources used within a given transcription and annotation campaign. Tool implementers and project managers are usually those who consider these specific aspects.

The second important aspect to consider is the ecology within which a given corpus creation project will take place and how much this may impact the issue of formats. In general, specific standards for representing a given transcription or annotation layer are chosen based on a wide variety of factors:

- In some cases, the choice will simply be dependant on the formats employed by the software used for the annotation task, and, to a lesser degree, how the tool exports data and files;
- The targeted representation format of an annotated corpus may depend on the kind of treatments that will be further operated upon the data. The capacity, for instance, of a query environment to have a more or less deep understanding of complex annotations or of combinations of various mark-up schemes will increase, or not, the actual requirements on the data formats;
- One has to consider which data structure the final corpus will be recorded in and archived in the long run. Indeed, combining too many heterogeneous formats, which might not all have the same level of stability and documentation, may hinder the further exploitation of the data outside (in time and space) the initial production locus;

- Finally, an important factor is the culture that a given community shares about standards and how difficult it is for community members (and groups of them) to change their practices. This learning curve effect usually explains why communities tend to design their own formats, to be able to progressively add layers of complexity.

## Basic components of an annotation schema

As explainedin the various contributions tothis book, each annotation tool tends to come with its own annotation schema and, in turn, each annotation schemais defined according to its own technical principles, mostly resulting from both legacy practices in the corresponding research environment and the actual preferences of the implementer. As a whole, it is seldom the case that an annotation schema results from a clear conceptual analysis where, in particular, the modelling (e.g. based on a UML specification) and representation (in the form of an XML schema for instance) levels are clearly differentiated (cf. Zipser and Romary, 2010). If we want, in this context, to move toward better interoperability across the existing initiatives within the spoken corpora community, it is necessary for us to introduce some basic elements that will act as references for comparing existing schemas and above all for mapping them onto common principles and standards.

The first stage for us is to define what is meant by an annotation and identify its various components. As illustrated in Figure 1, we consider that an annotation is a combination of three components, a source, a range and a qualifier, that have the following characteristics:

- The source[3] is the information upon which some additional statement is made in the context of the annotation. It is considered as a fixed object from the point of view of the annotation (i.e., changing the source invalidates the annotation);

---

[3] One may want to distinguish between a *primary source,* which is not anchored on any previous information layers and *secondary source*, when this can be seen as being derived from or built upon another source.

- The range identification characterises a portion of the source (a markable) that is being qualified by the annotation, either as one or several already identified parts of the source, or by reference to a certain identification scale (e.g. a temporal or spatial reference) that maps onto the source;
- The qualification expresses a constraint on the actual portion of the source as elicited by the range. This constraint is made of an elementary piece of information, mostly expressible as a feature-value pair.

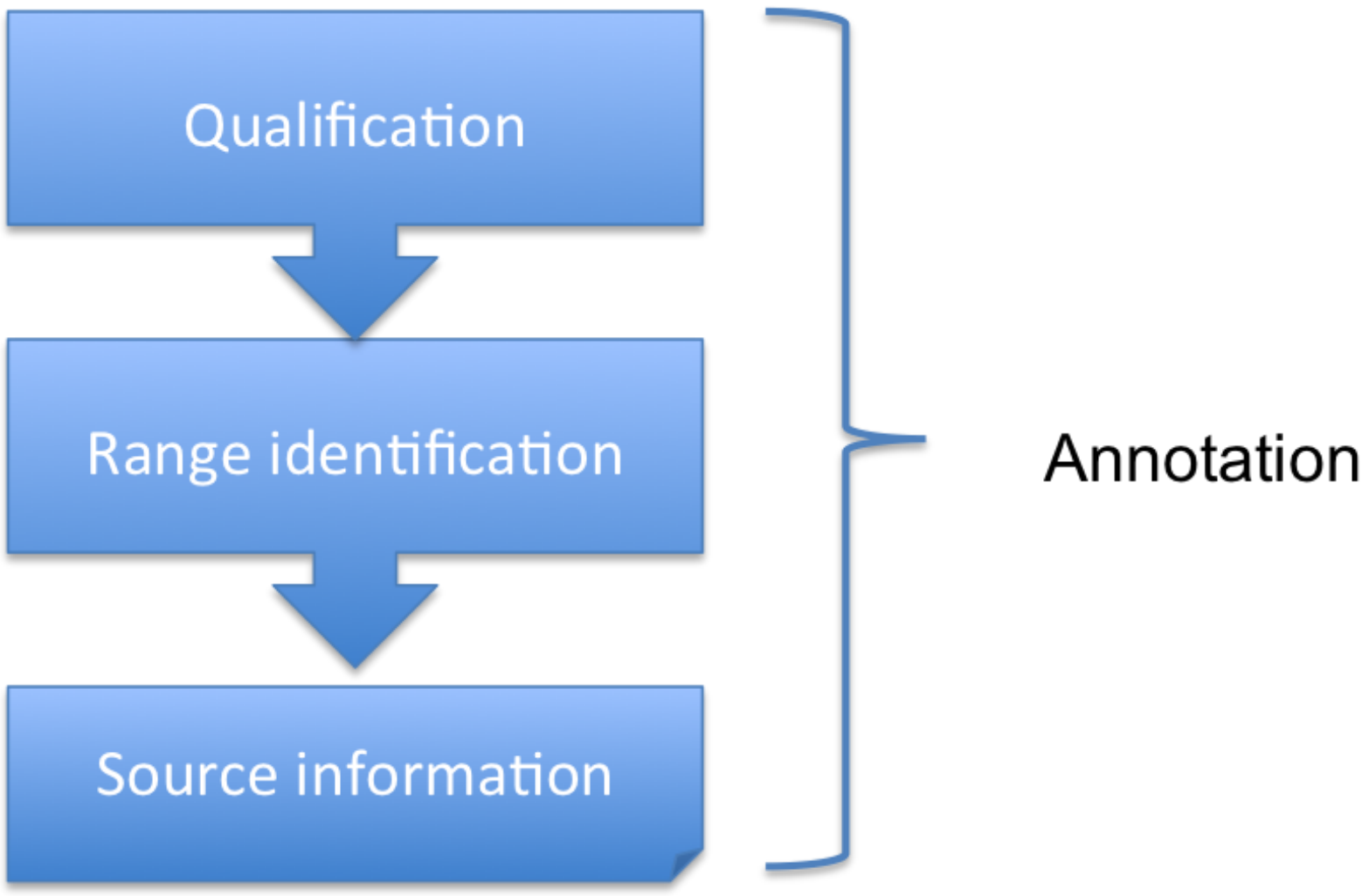

**Figure 1: generic structure of an annotation**

"Range" is defined hereabstractlybecauseexisting annotation schemas implement ranging mechanisms in different ways. These can be classified along the following lines:

- Direct reference to a (generally temporal) scale that transitively relates to the source. This is the basic mechanism provided by simple models such as annotation graph (Bird &Liberman, 2001). In such situations, there is no possibility to express an explicit co-occurrence relation between two annotations, except for identifying that one temporal reference is, for instance, the same. This is typically the strategy implemented in tools such as ANVIL or Praat (see Schmidt, 2011) ;
- Reference to reified objects on a scale. In the case of a temporal scale, this corresponds to the identification of events to which more than one qualifier may refer (i.e. a timeline, as in EXMARaLDA and ELAN, see Schmidt, 2011);

- Reference to explicit components of the source, allowing one to skip in some sense the actual ranging mechanism, or at least to make it boil down to a simple pointer or group of pointers. The important partof this last possibility is that it allows annotations to be about any kind of entity, including annotations themselves. This is usually the case for all annotation tools adopting a pure stand-off strategy such as MMAX (Müller and Strube, 2001).

An annotation is defined at a very low level of granularity, so thateach elementary statement upon a source (e.g. an elementary provision of some part of speech information about a word) is potentially embedded within a single annotation. Naturally, this does not preventspecific implementations fromproviding explicit factorisations that may facilitate the reduction of redundant information across annotations. For instance, a morpho-syntactic annotation schema may want to combine all information relevant toa given word by conflating all descriptors associated to a single range as a tagset label. Because there are so many types of linguistic annotation, annotations are often grouped according to different criteria. The reason for the grouping is technical and/or conceptual. To distinguish between these two different groupings, a distinction between annotation *layer* and annotation *level*has been introduced (see for exampleGoecke et al. 2010).A short description of this distinction can also be found in Witt (2004):

> To avoid confusion when talking about multiply structured text and text ideally organized by multiple hierarchies, the terms „level“ or „level of description“ is used when referring to a logical unit, e. g. visual document structure or logical text structure. When referring to a structure organizing the text technically in a hierarchically ordered way the terms „layer“ or „tier“ are used. A level can be expressed by means of one or more layers and a layer can include markup information on one or more levels.

Furthermore, it is possible to conceptualise the underlying coherence that is required when optimizing an annotation schema by defining the notion of annotation level as a coherent set of annotation types sharingthe following characteristics:

- Same underlying source, or set of sources (in the case of a corpus);
- Same ranging mechanism, by which we mean not only the same referring mechanisms (component or scale), but also a coherent description of

ranges from the point of view of their linearity, possible overlapping or alternation;

- Precisely defined and comprehensive data category selection that is applicable for qualifiers. We predict a general notion of tagset as such a selection.

With this general analysis in mind, we can now take a more precise look at the current state of standardisation processes forlanguage resources.

## Providing a reference semantics for linguistic annotation

One important aspect ofrepresenting anykind of annotation is the capacity to provide a clear and reliable semantics for the various descriptors that are being used, either in the form of features and featurevalues, or directly as objects in a representation expressed, for instance, in XML. In order to be shared across various annotation schemas and encoding applications, such a semantic should be implemented as a centralised registry of concepts, which we will henceforth refer to as data categories. As such, data categories should bear the following constraints:

- From a technical point of view, they must provide unique and stable references (implemented as persistent identifiers) such that the designer of a specific encoding schema can refer to them in his or her specification. By doing so, two annotations will be considered as equivalent when they are actually defined in relation to the same data categories (as feature and feature-value);
- From a descriptive point of view, each unique semantic reference should be associated with precise documentation combining a full text elicitation of the meaning of the descriptor with the expression of specific constraints that bear upon the category.

In recent years, ISO has developed a general framework for representing and maintaining such a registry of data categories, encompassing all domains of language resources. This work, carried out in the context of ISO project 12620 [ISO 12620], has led to the implementation of an online environment providing access to all data categories which have been standardised in the context of the various language resource-related activities within ISO, or specifically as part of the maintenance of the data category registry. It also provides access to the various data categories that individual language technology practitioners have defined in the course of their own work and decided to share with the community.

The ISO data category registry, as available through the ISOCat implementation, is meant to be a "flat" marketplace of semantic objects, providing only a limited set of ontological constraints. The objective there is to facilitate the maintenance of a comprehensive descriptive environment where new categories are easily inserted and reused without requiring any strong consistency check with the registry at large. Indeed, the following basic constraints are actually part of the data category model, as defined in ISO 12620:

- Simple generic-specific relations, when these are useful for the proper identification of interoperability descriptors between data categories. For instance, the fact that /properNoun/ is a sub-category of /noun/ allows one to compare morphosyntactic annotations which are based on different descriptive levels of granularity;
- Description of conceptual domains, in the sense of ISO 11179 ([ISO 11179]), to identify, when known or applicable, the possible value of so-called complex data categories[4]. For instance, this can be used to record that possible values of /grammaticalGender/ (limited to a small group of languages, see [Romary 2011]) could be a subset of {/masculine/, /feminine/ and /neutral/};
- Language-specific constraints, either in the form of specific application notes or as explicit restrictions bearing upon the conceptual domains of complex data categories. For instance, one could express explicitly that /grammaticalGender/ in French can only take the two values: {/masculine/ and /feminine/}.

In this section, we have tried to delineate a comprehensive view on annotations that, as it were, encompasses all types of representations within a multi-tier annotated corpus. Indeed, any kind of information added to a bare primary source (like an audio recording), from low-level segmentation markers to high-level discourse relation identification, can be seen as an annotation in the sense presented here.

---

[4] *Complex* data categories will typically be implemented as place-holders (or features), whereas *simple* data categories, will be implemented as values.

# Language resource management – an ISO perspective

## Specific ISO models and formats for linguistic annotation

ISO committee TC 37/SC 4, launched in 2002, focuses on the definition of models and formats for the representation of annotated language resources. To this end, ISO/TC 37/SC 4 has generalised the modelling strategy initiated by its sister committee SC 3 for the representation of terminological data [Romary, 2001], and through which linguistic data models are seen as the combination of a generic data pattern (a *meta-model*), which is further refined through a selection of *data categories*, which provide the descriptors for this specific annotation level. Such models are defined more or less independently from any specific formats (not even bound to an XML framework), and ensure that an implementer has the necessary tool to design and compare formats, with regard to their degrees of interoperability. In the rest of this section, we will go through several projects[5] from ISO/TC 37/SC 4 that are important for phonological corpora.

One of the early proposals of ISO/TC 37/SC 4 has been to outline a possible standard for morpho-syntactic (also referred to as *part-of-speech*) annotation. Such an annotation level corresponds more or less to the first linguistic abstraction level for a corpus and, depending on the language to be annotated and the actual characteristics of the tool that is being used, can vary enormously in structure and complexity. In order to deal with the complex issues of ambiguity and determinism in morpho-syntactic annotation, ISO 24611/MAF makes a clear distinction between the two levels of *token*s (representing the surface segmentation of the source) and *word forms* (identifying lexical abstractions associated to groups of tokens). These two levels have the specificities that, on the one hand, they can be represented as simple sequences as well as local graphs (e.g. multiple segmentations, ambiguous compounds, etc.), and, on the other hand, any n to n combination can stand between word forms and tokens[6]. Associated to this meta-model, MAF provides a default XML syntax, but as we shall see later in this chapter, it is also possible to contemplate a TEI-based implementation for it.

[5] In the ISO sense

For syntactic annotation, however, ISO committee TC 37/SC 4 did not reach an early consensus on a possible XML syntax that would cover the variety of possible syntactic frameworks (constituency or dependency based, theory specific) that can be observed either within existing treebanks (Abeillé, 2003) or as export formats of syntactic parsers (Ide&Romary, 2003). The published standard (SynAF, ISO 24615) is thus centred on a comprehensive meta-model informing the whole spectrum of syntactic representation practices, coupled with an extensive list of data categories that are now available within ISOCat (see section (Broeder et al., 2008). The standard can presently be used to specify new formats or make interoperability checks[7], and a reference serialisation of SynAF that would cover the kind of features now available in such formats as Tiger2 (Romary et al., preprint)[8] is planned.

Work carried out within ISO project 24617-2 provided a comprehensive framework for the annotation of dialogue acts [Bunt et al., 2010], applicable to any kind of multimodal interaction. ISO/DIS 24617-2 (Dialogue acts) can be seen at various levels of abstraction. It first provides a well-defined theoretical framework where the basic concepts of *dialogue act*, *semantic content* and *communicative function* are defined. Building upon the numerous initiatives and projects[9] that have taken place in the last twenty years, it defines a domain-independent meta-model providing a multidimensional description of dialogue act phenomena, coupled with data categories registered in the ISOCat registry.

[6] One token can correspond to several word forms, and vice versa.

[7] Usually to assess the conformity of a data set with an expected input of a tool, and design a possible filter accordingly.

[8] See the recent proposals by Głowińska and Przepiórkowski (2010) and Erjavec et al. (2010) for the encoding of SynAF compliant annotations by means of the TEI framework.

[9] Cf. annotation schemes defined in such projects as TRAINS, HCRC Map Task, Verbmobil, DIT, SPAAC, C-Star, MUMIN, MRDA, AMI, and more recent attempts towards domain-independence, interoperability and standardization in DAMSL, MATE, DIT++ or the EU project LIRICS.

Finally it offers a default XML serialisation that fully implements the features of the intended model[10].

As the preceding examples make clear, the focus on modelling and interoperability issues facilitates the design of a given corpus as the combination of basic standardisation building blocks, which can then be adapted by projects to handle legacy data or tools. It also allows one to anticipate possible transitions to make existing data more and more compliant to international standards when they are adopted within a scholarly community.

## Genericity made a principle: LAF – GRAF

In cases where no standardisation activity for a specific annotation level exists, or, as is usually the case, when a variety of annotation levels have to be merged within one single information pool in order to carry out cross-level queries or visualisation, there is a need for a high level representation that basically unifies all types of specific annotation structures. Various proposals have been suggested to addressthis situation, including projects such as ATLAS (Bird et al., 2000), Mate (McKelvie et al., 2001) or more recently the American National Corpus (ANC; Ide and Macleod, 2001). The American National Corpus project was an opportunity to experiment and finalise the principles enunciated in the ISO LAF project, on the basis of a generic graph representation where nodes represent the reification of linguistic annotation components and edges relations between them. Based on the ISO-TEI feature structure standard for the further qualification of nodes and edges, LAF offers a default format (called GraF) for the serialisation of any type of linguistic structure.LAF wascreatedto provide easy mapping with similar past and present initiatives such as annotation graphs, or PAULA. It is also an important step in contemplating generic query mechanisms and perhaps a standardised query language for language resources.

## Linguistic annotation with the TEI

In many respects, the TEI appears to be a most appropriate method to a) describe primary transcription of phonological corpora and b) implement the models provided by ISO standards (Romary, 2009). Indeed, the Text Encoding

[10] Even if space prevents us from providing further details on this, this serialization is inspired from the annotation framework provided by the TEI guidelines.

Initiative can be a good entry point for anyone looking for a generalpurpose XML vocabulary, which in turn may be connected to —and thus made interoperable with — many other corpora and encoding initiatives.

In the rest of the paper, we show how the TEI guidelines already offer a variety of constructs and mechanisms to cope with many issues relevant to spoken corpora and their annotations. When applicable, we will make the necessary links with ongoing ISO/TC 37 activities so that some clues are given as to how a possible transition to more elaborate annotation schemas, or possibly a mapping from basic TEI representations to other annotation schemas, could be implemented.

## The TEI framework for transcribing spoken corpora

The Text Encoding Initiative (TEI)beganin the late 1980s to propose approaches to annotate different types of textually represented resources. Beginning with the 3rd major edition of the TEI Guidelines (Sperberg-McQueen &Burnard, 1994), the TEI also addresses the topic of annotating transcribed speech. After a revision of the Guidelines in 2002 that mainly switched from an SGML- to a fully XML-compliant syntax of the annotation, the most recent version of the TEI-annotation scheme was published as TEI P5 in 2009 as a "living document" that is continuously updated. This section describes TEI's approach totranscribing spoken language according to P5 (TEI 2011). However, as the TEI consortium has been very careful with their updates and changes – especially the chapter on the transcription of spoken languages, which has only seen a few minor changes over the years – older TEI-based annotations are still usable without much effort. The general structure of the TEI encoding framework is highly modularised. About 30 specialised TEI modules exist, for instance for dictionaries, verse text, dramas, linguistic analysis, and speech transcriptions. Moreover, it is also possible to define freely specialised tag sets for all purposes not addressed by existing TEI tags.

Independent from the type of the annotated document, i.e. regardless of the used TEI modules, all TEI documents are subdivided into two major parts: the TEI-Header containing the metadata of the annotated resource, for instance information on the time and place a dialogue took place; and the annotated resource itself, for instance the transcription of the spoken dialogue. (see listing)

```
<TEI>
<teiHeader>
<!--Comment: -->
<!-- the metadata of the annotated resource are included here  -->
</teiHeader>
<text>
<!--Comment: -->
<!-- the annotated resource itself is included here  -->
</text>
</TEI>
```

The following sections describe the TEI-metadata and TEI-annotations with a strong focus on options to deal with spoken language. This entails the omissionof many aspects of TEI. The complete guidelines with its some 1300 pages are available on the TEI website (http://www.tei-c.org).

## The TEI Header

The header of the TEI document contains all the metadata associated with a spoken text. This information is subdivided into four different major classes: (1) the file description, (2) the encoding description, (3) the profile description, and (4) the revision description. While the revision description does not contain information specifically relevant tophonological resources, the other three do. Apart from the file description, all other parts of the header can be omitted.

```
<teiHeader>
<fileDesc>
<!-- ... -->
</fileDesc>
<encodingDesc>
<!-- ... -->
</encodingDesc>
<profileDesc>
<!-- ... -->
</profileDesc>
<revisionDesc>
<!-- ... -->
</revisionDesc>
</teiHeader>
```

### Information about the file

There are only three necessaryparts to a TEI Header. All of them must be included as children of the file description, annotated as <fileDesc>. These necessaryelements are used to provide information about the title (<titleStmt>), a publication statement (<publicationStmt>), and a description of the source of the annotated text (<sourceDesc>). In some respects, the file description contains information usuallyregarded as metadata. In case of annotated speech resources, this class also allows the representation of information about the source of the transcription, almost always a recording. Technical data of a speech recording can be included in the information contained in<sourceDesc>. Such data include file format information (e.g. uncompressed wav, compressed mp3 or ogg, the sampling frequency), specifications of the audio equipment (e.g. the number and the type(s) of microphone(s)), the source of the recording (e.g. original recording, broadcast transmission), etc. For this kind of information the <recordingStmt> (recording statement) with its sub-element <recording> (recording event) are available in the header of a TEI document that contains the transcription of speech.

```
<fileDesc>
<!-- ... -->
<sourceDesc>
<!-- ... -->
<recordingStmt>
<recording type="audio">
      <equipment>
      <p>Two microphones, standard 44.1 KHz sampling frequency</p>
</equipment>
<date>12 Jan 2010</date>
</recording>
</recordingStmt>
<!-- ... -->
</sourceDesc>
<!-- ... -->
</fileDesc>
```

The type of recording could also be ‘video’. In addition tothe description of the <equipment> used to prepare the <recording>, the element <broadcast> could be used if the source wererecorded from radio or TV. Of course, since the broadcast speech was also recorded before transmission, it is possible to include the element <recording> in <broadcast>, as well. This exemplifies how rich the TEI’s metadata description can be whenneeded.

### Information about the encoding

The encoding declaration “documents the relationship between an electronic text and the source or sources from which it was derived” (TEI P5). Besides other information the element <encodingDesc> allows a tagging declaration to provide detailed information about the tagset used in the document, the feature system declaration <fsdDecl> that could be used when applying feature structures, and the element <geoDecl> for the declaration of the geographic coordinates.

Because a lot of transcriptions of spoken language are prepared (semi-)automatically, for instance with the tools described in this volume, one might want to mention which tools have been used for this task in the metadata. The element <appInfo> allows the specification of a list of applications used for preparing the transcription.

```
<appInfo>
<application version="1.4.4" ident="EXMARaLDA">
<label>EXMARaLDAPartitur-Editor</label>
<ptr target="#dialog2"/><ptr target="#dialog132"/>
</application>
</appInfo>
```

This example defines the application EXMARaLDAPartitur-Editor 1.4.4 and specifies two dialogues that have been transcribed with this tool.

## Information about the profile

A comprehensive description of the languages used by the speakers, information about the situation in which the speech recording took place and other non-bibliographic metadata can be specified in a profile description.

One important component for the transcription of speech, especially when elicited in an experiment, is the <settingDesc>. By means of this element it is possible to provide information about the place, date, activities etc. of the speech interaction. It could also be used to refer to controlled settings as e.g. in Maptask-(Anderson et al. 1991) and Tinkertoy(Senft 1994) experiments.

It is possible to provide very fine-grained metadata with very detailed specifications of a participant in a dialogue. Within the <profileDesc> the element <partDesc> can be used to include information about participants in a conversation by means of a list of <person>-elements. This element enablespersonal data for a person to be included, for instance:

```
<person sex="2" age="infant">
<birth when="2010">
<date>12 Jan 2010</date>
<name type="place">Berlin, Germany</name>
</birth>
<langKnowledge tags="de ">
<langKnown level="first" tag="de">German</langKnown>
</langKnowledge>
</person>
```

## TEI-based transcription

In this section, we discusshow the TEI can be used for spoken data, using a dialogue from Thomas Schmidt's article in this volume as an example. In this example the persons communicate verbally in French as well as through gestures. A translation into English and additional information are also provided. Furthermore, the alignment of the characters and the timeline indicate the sequence and the overlap of information.

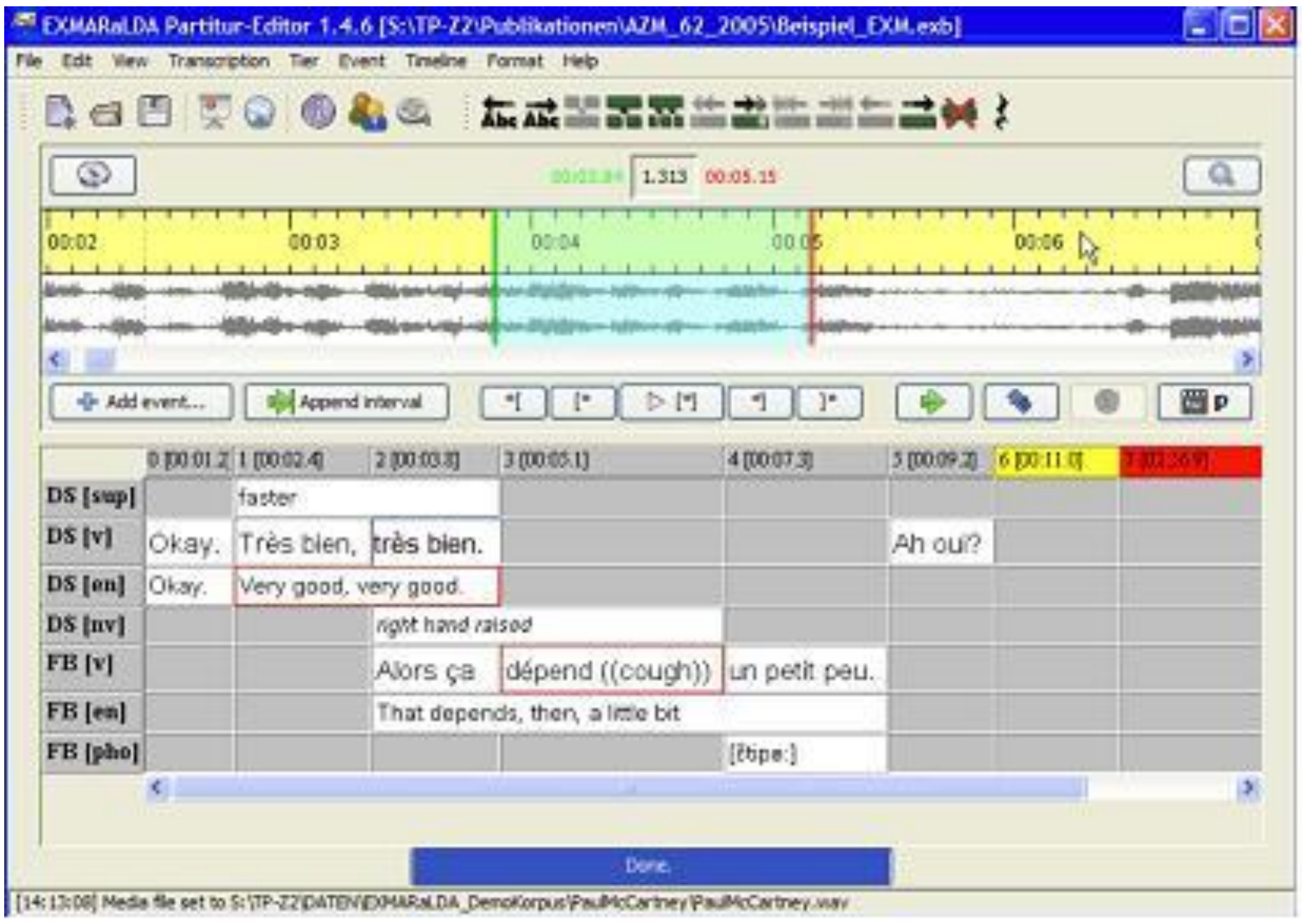

Figure 2: Example of a transcription using EXMARalDa

Whereas the metadata of a speech transcription are embedded in the <teiHeader>, the actual transcriptions are part of the <body> of a TEI document. The <body> embeds one or more 'utterances' (<u>). Within an element <u> an orthographic or a phonetic transcription is included. Since this element may contain text, it is possible to include annotations in non-XML-based conventions. The following example uses the convention GAT (see Schmidt, this volume) to mark a non-linguistic event.

```
<u>Alorsçadépend ((cough))un petit peu.</u>
```

Such an approach allows researchers to continue to use conventions they are used to. At least, they can do so to a certain extent, as long as the annotation conventions do not contradict constraints pertaining to text data in XML documents. This means, in particular, that characters like '<' or '&' cannot be included directly, instead they have to be represented as so-called XML entities.

As a result of this restriction, the widely used CHAT conventions (REF) cannot be included here directly.For example, the event in the samplesentence above would be annotated as `&=coughs` in CHAT syntax. But even in cases that do not lead to such difficulties, it is not recommended to mix syntactic variants. The TEI tagset for transcriptions of spoken language defines several elements for the integration of annotation. The TEI-conformant representation of the utterance given above would be:

```
<u>Alorsçadépend<vocal><desc>cough</desc></vocal> un petit peu.</u>
```

In general, the element `<vocal>`should be used for non- or semi-lexicalised sounds. Other elements, like `<kinesic>` and `<incident>` could be used to mark gestures, environmental noise, etc.:

```
<kinesic><desc>right hand raised</desc></kinesic>
```

Because XML documents are, technically, nothing but a sequence of characters, indentation and visual alignment are not usable in order to indicate relations like the synchronicity or overlap of utterances, gestures, occurrences in the environment etc. Instead of visual alignment, XML enforces the use of special mark up in order to make such relations explicit. This can be done according to varying degrees of detail. On the one hand, we can mark up information corresponding to simple statements like "a speaker started an utterance before the other speaker finished her utterance". On the other hand, we may have something like an explicit reference to a timeline. The following example shows an approach whose grade of granularity ranges between these two extremes:

```
<body>
<u who="#SPK1">Okay. Trèsbien,
<anchorxml:id="tp1u"/>trèsbien.<anchor xml:id="tp2u"/></u>
<u who="#SPK2"><anchor synch="#tp1u"/>Alorsçadépend
<vocal><desc>cough</desc></vocal>
<anchorxml:id="tp2u"/>un petit peu.</u>
<kinesic who="SPK1" type="nv" start="#tp1u">
<desc>right hand raised</desc></kinesic><anchor synch="#tp2u"/>
<u who="#SPK1">Ah oui?.</u>
</body>
```

In this example, the overlapping speech of the two speakers is indicated by the inclusion of an anchor within the first utterance at the point where the second speaker starts his or her first utterance. At this very point the first speaker starts a gesture that ends when the second speaker begins the phrase “un petit peu”. Besides this explicit information about the temporal relations of the different utterances and gestures, implicit temporal information is also included in the XMLfile, simply due to the serialisation of the XML document. If there is no explicit information about overlaps, then it is implied that the communication events (speech, gestures etc.) have been produced sequentially one after the other. In the example above, this means that the last utterance ‘Ah oui?.’ starts after the completion of its previous speech turn “Alorsça depend ((cough)) un petit peu.” The most precise approach to keep the temporal information is referencing each event to relative or absolute time points. This can be done by including the TEI element <timeline>, the definition of relevant time points and linking from utterances etc. to them. In the annex of this chapter a complete example that makes use of this technique is given.

One of the most interesting benefits when using a TEI-based approach to annotate speech corpora is the possibility of including elements from all other TEI modules. One of these modules is described in the TEI guidelines in chapter 17,“Linking, Segmentation, and Alignment”. It not only provides elements for a highly sophisticated addressing and linking mechanism, but also an element <seg> that allows the grouping of text fragments as long as the XML constraints are met. So, naturally, it is not possible to split elements in a way that results in overlapping markup. The element <seg> might be used with the attribute ‘xml:id’ to provide unique identifiers. This allows, whenever needed, a direct referencing to arbitrary text segments. Another example of the use of the <seg>element given in the TEI Guidelines (TEI 2011, pp. 464f.) is reproduced below:

```
<seg type="sentence" subtype="declarative">
<seg type="phrase" subtype="noun">
<seg type="word" subtype="adjective">Literate</seg>
<seg type="word" subtype="conjunction">and</seg>
<seg type="word" subtype="adjective">illiterate</seg>
<seg type="word" subtype="noun">speech</seg>
</seg>
<seg type="phrase" subtype="preposition">
<seg type="word" subtype="preposition">in</seg>
<seg type="word" subtype="article">a</seg>
<seg type="word" subtype="noun">language</seg>
<seg type="word" subtype="preposition">like</seg>
<seg type="word" subtype="noun">English</seg>
</seg>
<seg type="phrase" subtype="verb">
<seg type="word" subtype="verb">are</seg>
<seg type="word" subtype="adverb">plainly</seg>
<seg type="word" subtype="adjective">different</seg>
</seg>
<seg type="punct">.</seg>
```

```
</seg>
```

In this example the element <seg> is used to segment a sentence into phrases and words and to associate more detailed information like the phrase type or the part of speech with the segments. However, the guidelines also make clear that a more appropriate annotation of linguistic information is available in the module "Simple Analytic Mechanisms", because this module defines not only specialised elements for sentences (<s>), phrases (<phr>) and words (<w>) but also for morphemes (<m>) and syllables (<syll>).

## Annotating corpora with the core mechanisms of the TEI

### Using feature structures within an annotation scheme

In this section we address the implementation of what we named the'qualification level' by means of feature structures and compare it with the general model for elementary annotations describedabove. Feature structures (Pollard & Sag, 1987) are formal structures which combine a basic representation mechanism by means of a possibly recursive combination of feature-value pairs, where values can in turn be feature structures and associated operations in order to access, filter or unify such structures. Feature structures have been used as the reference mechanism for various unification-based formalisms andalso as a descriptive tool in order to attach basic properties to a linguistic segment (e.g. for phonetic descriptions, [Bird & Klein, 1994]). Complementing this well-established scientific background, an XML-based representation for feature structures has been developed since the early days of the Text Encoding Initiative by Terry Langendoen and Gary Simons (1995), and has been further improved and stabilised in the context of a joint TEI-ISO activity (ISO 24610-1, henceforth ISO-TEI-FSR).

The representation of feature structures in ISO-TEI-FSR is based upon two central elements:

- `<f>` which contains a single feature-value pair
- `<fs>` which groups together one or several feature-value pairs

A simple feature-value pair is described by means of the name of the feature (attribute @name) and its value, expressed as the content of the <f> element. In the canonical ISO-TEI-FSR this value is systematically typed by means of an embedded element, which can either be `<binary>` (with attribute @value=true/false), `<symbol>`, `<numeric>` or `<string>`.

For instance, the expression of a part of speech value for a noun would typically look like this:

```
<f name="partOfSpeech">
<symbol>noun</symbol>
</f>
```

When combined, several feature-value pairs should be embedded within a feature structure, which can optionally be further typed, for instance to provide direct access to all feature structures associated withthe same annotation level. For instance, a basic morphosyntactic qualification block could be represented as:

```
<fs type="morphosyntacticAnnotation">
	<f name="partOfSpeech">
		<symbol>noun</symbol>
	</f>
	<f name="grammaticalGender">
		<symbol>masculine</symbol>
	</f>
	<f name="grammaticalNumber">
		<symbol>plural</symbol>
	</f>
</fs>
```

As an illustration of the way feature structures can be used to describe the basic components of an annotation scheme, let us show how atagset can be covered with this framework.

## Creating tagsetsthrough feature structure libraries

### Rationale

The main issue regardingtagsets[11] as reference descriptions for morphosyntactic annotations is that they can be shared across corpora and annotation tools. In particular, a tagset articulates the relation between a concrete syntactic representation within a set of annotationsanda reference semantics that may allow one to interpret the annotation further when exploring the annotated data. To this end, the ISO-TEI standard provides mechanisms for declaring feature and feature-value libraries that perfectly match the objective stated here.

[11] See for instance (Monachini and Calzolari, 1994) for the corresponding work carried out within the Multext project.

In the following section we will briefly outline a possible method for declaring tagsets in the feature structure framework, to show that such a method could be used as reference to actually document, record and compare the various tagsets used within the linguistic and computational linguistic communities.

### Description of an elementary tag

The first step in the process of declaring a tagset is the ability to describe elementary features. This can easily be achievedwiththe ISO-TEI standard by combining elementary feature statements such as those seen above within a feature library (fLib), together with a systematic identification of each feature (by means of an *xml:id* attribute).

In the following example, the three elementary features corresponding to the grammatical gender possibilities in German are described accordingly.

```
<fLib n="grammatical gender" >
<f name="grammaticalGender" xml:id="fem">
<symbol value="feminine"/>
</f>
<f name="grammaticalGender" xml:id="mas">
<symbol value="masculine"/>
</f>
<f name="grammaticalGender" xml:id="neu">
<symbol value="neuter"/>
</f>
</fLib>
```

It can be noted here that if desired, one may fragment the various types of features (grammatical category, gender, number etc.) within separate `<fLib>` constructs or just group them all together within a single one. For instance, and in order to have all the illustrative material at hand, we could have the following series of declarations for grammatical categories:

```
<fLib n="grammatical category">
<f name="partOfSPeech">
<symbol value="commonNoun" xml:id="#NC"/>
</f>
<!-- further grammatical categories here -->
</fLib>
```

as well as for grammatical number:

```
<fLib n="grammatical number">
<f name="grammaticalNumber">
<symbol value="singular" xml:id="sing"/>
</f>
<!-- further values for grammatical number here -->
</fLib>
```

### Description of a complete tagset

Once all the elementary declarations are made, the ISO-TEI framework allows one to combine them to declare feature-value libraries (fvLib), within which a feature structure combining elementary morpho-syntactic features corresponds to a tag in the tagset in a one-to-one manner. In the following (simplified) example, for instance, the tag for a masculine singular common noun is declared and provides the appropriate identifier for further reference:

```
<fvLib>
<fsxml:id="Ncms__" feats="#NC #mas #sing"/>
<!-- further tags declared here -->
</fvLib>
```

Once such a full tagset is described, the various entries maybe reused in many different ways. In a proprietary format, it may simply be referred to in the documentation in order to provide a formal reference to the corresponding annotation scheme, or, when available, it can be referred to within the declaration section of an annotation file. In the case of a fully TEI-based representation, a possible mechanism is to see a tag as an analysis of a linguistic segment and point to the declaration by means of the @ana attribute, as in the following example[12]:

```
<p><w>Le</w><w>petit</w><w
ana="#Ncms__">chat</w><w>est</w><w>mort</w><pc>.</pc></p>
```

### Towards maintainable and sustainable specifications

The standard-based description of tagsets outlined above only makes sense if the actual specifications can actually be re-used as a reliable reference across various annotation projects. Even in basic cases, where such feature-structure libraries can be imbedded within the document containing the data itself (like in the <back> component of the TEI document), this is obviously not a good strategy if one wants to maintain and disseminate a tagset specification in a sustainable way. It is thus a recommended best practice to integrate tagset specifications within their own TEI document, which in turn lets one document and record origin and versioning information in the corresponding TEI header.

Once one has a stable tagset specification at hand, it is probably time to consider a dissemination and standardisation strategy. First, we recommend storing the specification in a stable registry, with version control mechanisms (such as SourceForge). This can be a way to involve a wider community in using and reacting to the proposed tagset. The second stage is to build a real standardisation strategy, either by making the tagset a recommendation of an institution or a research infrastructure (such as CLARINor DARIAH), or by actually making this a contribution to ISO/TC 37/SC 4 (as a technical report, for instance). It should be noted here that any such move toward a wider publication of a specification will result in requests for evolution.

---

[12] Molière, *L'école des femmes*, II(5), 461.

A final word on the issues of publication, dissemination and above all standardisation: we recognise the need for several reference tagsets for a given language. Depending on the use case, or the expected granularity of description, tagsets may vary in the way they use and combine morpho-syntactic features. Still, the proper publication — in a standardised format, as suggested here — of the tagset specification, as well as its systematic anchoring tothe data categories in ISOCat, will improve our capacity to provide better comparisons between them.

## Range identification in the TEI framework

To complement the use of feature structures that we presented above, the TEI provides mechanisms for the annotation of ranges and their linking to qualifications (as described in Figure 1). In the following section we will briefly describe this mechanism in order to provide a comprehensive package for linguistic annotation.

The central element for range identification in the TEI guidelines is <span>, which specifies, by means of a `@from` and a `@to` attribute, a sequence within a document to which ones want to make an annotation. In its simplest form, <span> allows one to make a plain text comment in the element content. In the case of formal annotations, `<span>` bear the `@ana` attribute that we have already seen to point to a structured qualification such as a feature structure. Furthermore, the TEI provides a `<spanGrp>` element to put together all span descriptions that correspond, for instance, to the same annotation level.

To illustrate the possible use of the `<span>` element in a concrete annotation case, let us consider the morpho-syntactic annotation of a linguistic sequence in conformance with the ISO MAF proposal. By construction, the MAF meta-model makes a clear (and essential) distinction between a token level and a word form level. The token level corresponds to the identification of elementary segments on the linguistic surface, whereas word forms are abstract lexical items identified across spans of one or several tokens. This model can be implemented within a full TEI-based representation by means of `<span>` as follows: the transcription is initially tokenised by means of the `<w>` element, as presented before. We take here a simple sequence ("pomme de terre", *potato*) corresponding to a compound lexical item:

```
<uwho="#speakerA">
   ...
<w xml:id="t1">pomme</w>
<w xml:id="t2">de</w>
<w xml:id="t3">terre</w>
...
</u>
```

The word form level is then implemented by means of `<span>`s that can be set together within a single <spanGrp>:

```
<spanGrp type="wordForm">
   ...
<span from="#t1" to="#t3" ana="#pomme_de_terre_sing"/>
...
</spanGrp>
```

Each <span> is actually pointing to a reference lexical entry and more precisely to the corresponding inflected form. Such a lexical entry can be implemented, in compliance with ISO LMF, as a TEI `<entry>` element, as follows (excerpt):

```
<entry>
<form type="inflected" xml:id="pomme_de_terre_sing">
<orth>pomme de terre</orth>
<gramGrp>
<number>singular</number>
</gramGrp>
</form>
<form>
...
</form>
</entry>
```

## Conclusion — further standard developments in the domain of spoken corpora

In roughly the last 25 years, pioneering work has laid the groundwork for a wide-coverage standardisation framework, which, combining the existing background from both the TEI and ISO, offers a wide range of possibilities to deal with both primary transcriptions and higher level annotations. In this paper we hope we have conveyed the message that, within what could appear as an intricate jungle of standards, it is possible to identify some baseline formats allowing one to start putting together a corpus project within some stable normative environments such as the TEI. By doing so, we also want to suggest that the phonological corpora community should become less and less dependent upon proprietary formats created for specific projects and design its own standardisation roadmap to both improve existing proposals and fill in the gaps that still exist in this domain (e.g. representation of multiple tiers within an annotated corpus). This should be accompanied by a stronger involvement by the spoken corpus community in standardisation bodies such as ISO or the TEI, as well as more effortstoward the identification and dissemination of an optimal combination of standardswhich could be delivered as guidelines of best practices to the community.

## Annex

The following XML excerpt represents a fully compliant example of a TEI-based representation for a transcription of speech. It illustrates in particular the use of a timeline mechanism to anchor the transcription toreference temporal points in the source audio file.

```
<TEI xmlns="http://www.tei-c.org/ns/1.0">
<teiHeader>
<fileDesc>
<titleStmt>
<title>Title</title>
</titleStmt>
<publicationStmt>
<p>Publication Information</p>
</publicationStmt>
<sourceDesc>
<p>Information about the source</p>
</sourceDesc>
</fileDesc>
<profileDesc>
<particDesc>
<personxml:id="SPK0">
<persName>
<abbr>Peter Black</abbr>
</persName>
</person>
<personxml:id="SPK1">
<persName>
<abbr>Judith White</abbr>
```

```
</persName>
</person>
</particDesc>
</profileDesc>
</teiHeader>
<text>
<timeline unit="ms">
<whenxml:id="T1"/>
      <whenxml:id="T2"/>
<whenxml:id="T3"/>
<whenxml:id="T4"/>
<whenxml:id="T4bar"/>
<whenxml:id="T5"/>
<whenxml:id="T6"/>
<whenxml:id="T7"/>
</timeline>
<body>
      <u who="#SPK0"><anchor synch="#T1"/>Okay. <anchor
synch="#T2"/>Très bien, <anchor synch="#T3"/>très bien.<anchor
synch="#T4"/></u>
<u who="#SPK1"><anchor synch="#T3"/>Alors ça<anchor
synch="#T4"/>depend <anchor synch="#T4bar"/><kinesic
type="cough"/><anchor synch="#T5"/>un petit peu. <anchor
synch="#T6"/></u>
<incident who="SPK0" type="nv" start="T3" end="T5">
<desc>right hand raised</desc>
</incident>
<u who="#SPK0"><anchor synch="#T6"/>Ah oui?. <anchor
synch="#T7"/></u>
</body>
</text>
</TEI>
```